\documentclass{article}

    \usepackage[preprint]{neurips_2025}

\usepackage[utf8]{inputenc} %
\usepackage[T1]{fontenc}    %
\usepackage{hyperref}       %
\usepackage{url}            %
\usepackage{booktabs}       %
\usepackage{amsfonts}       %
\usepackage{nicefrac}       %
\usepackage{microtype}      %
\usepackage{xcolor}         %

\usepackage{adjustbox}
\usepackage{amsmath}
\usepackage{amssymb}
\usepackage{mathtools}
\usepackage{amsthm}
\usepackage{multirow}
\usepackage{adjustbox}
\usepackage{colortbl}
\usepackage{booktabs} %
\usepackage{url}
\usepackage{amssymb}
\usepackage{bbding}
\usepackage{pifont}
\usepackage{wasysym}
\usepackage{utfsym}
\usepackage{fontawesome}
\usepackage{natbib}
\setcitestyle{numbers,square}
\usepackage{hyperref}
\hypersetup{
    colorlinks=true,
    linkcolor=blue,  
    urlcolor=cyan,
    }
\usepackage{arydshln}
\usepackage{array}

\title{An Empirical Study of Qwen3 Quantization}

\author{Xingyu Zheng\textsuperscript{*}$^{1}$, Yuye Li\textsuperscript{*}$^{2}$, Haoran Chu\textsuperscript{*}$^{1}$, Yue Feng\textsuperscript{*}$^{1}$, Xudong Ma$^{1}$, \\
\textbf{Jie Luo}$^{1}$ \textbf{, Jinyang Guo}$^{1}$ \textbf{, Haotong~Qin}\textsuperscript{\dag}$^{3}$ \textbf{, Michele Magno}$^{3}$ \textbf{, Xianglong Liu}$^{1}$\\
$^{1}$Beihang University\quad
$^{2}$Xidian University\quad
$^{3}$ETH Z\"{u}rich\\
\texttt{\small\{zhengxingyu,23371505chr,fay777,macaronlin,luojie,jinyangguo,xlliu\}@buaa.edu.cn}\\
\texttt{\small liyueye541@gmail.com \small\{haotong.qin,michele.magno\}@pbl.ee.ethz.ch}
}

\begin{document}

\maketitle

\begingroup
\renewcommand\thefootnote{\fnsymbol{footnote}}
\footnotetext[1]{Equal Contribution. $^\dag$ Corresponding Author.}
\endgroup

\vspace{-0.10in}
\begin{abstract}
The Qwen series has emerged as a leading family of open-source Large Language Models (LLMs), demonstrating remarkable capabilities in natural language understanding tasks. With the recent release of Qwen3, which exhibits superior performance across diverse benchmarks, there is growing interest in deploying these models efficiently in resource-constrained environments. Low-bit quantization presents a promising solution, yet its impact on Qwen3's performance remains underexplored. This study conducts a systematic evaluation of Qwen3's robustness under various quantization settings, aiming to uncover both opportunities and challenges in compressing this state-of-the-art model.
We rigorously assess 5 existing classic post-training quantization techniques applied to Qwen3, spanning bit-widths from 1 to 8 bits, and evaluate their effectiveness across multiple datasets. Our findings reveal that while Qwen3 maintains competitive performance at moderate bit-widths, it experiences notable degradation in linguistic tasks under ultra-low precision, underscoring the persistent hurdles in LLM compression. These results emphasize the need for further research to mitigate performance loss in extreme quantization scenarios.
We anticipate that this empirical analysis will provide actionable insights for advancing quantization methods tailored to Qwen3 and future LLMs, ultimately enhancing their practicality without compromising accuracy. Our project is released on \href{https://github.com/Efficient-ML/Qwen3-Quantization}{GitHub} and \href{https://huggingface.co/collections/Efficient-ML/qwen3-quantization-68164450decb1c868788cb2b}{Hugging Face}.
\vspace{-0.10in}
   
\end{abstract}

\section{Introduction}
Developed by Alibaba Group, the Qwen series~\cite{bai2023qwen,yang2024qwen2} has rapidly advanced as a competitive open-source family of autoregressive large language models (LLMs) based on the Transformer architecture~\cite{vaswani2017attention}. With its initial release in 2023, Qwen demonstrated exceptional scalability, with even its 7B parameter model rivaling larger proprietary models like GPT-3.5 in certain benchmarks. The recently launched Qwen3\footnote{\ \url{https://github.com/QwenLM/Qwen3}\label{foot:qwen3}}, available in configurations from 0.6B to 235B parameters, further elevates performance through refined pre-training on diverse, high-quality corpora. This positions the Qwen family among the most capable open-source LLMs, adaptable to diverse deployment scenarios.

Despite its strengths, practical deployment of Qwen3 faces challenges due to high computational and memory demands. Low-bit quantization~\cite{xiao2023smoothquant,frantar2022gptq,huang2024empirical,gong2024survey} has emerged as a critical technique to mitigate these issues, enabling efficient inference on resource-constrained devices. However, quantization often introduces performance degradation. Qwen3’s state-of-the-art capabilities present a timely opportunity to reassess quantization techniques, uncovering new insights into their efficacy and limitations for cutting-edge models.

In this empirical study, we systematically evaluate Qwen3’s robustness under quantization across Post-Training Quantization (PTQ) methods. We test 5 classic methods, including Round-To-Nearest (RTN), GPTQ~\cite{frantar2022gptq}, AWQ~\cite{lin2023awq}, SmoothQuant~\cite{xiao2023smoothquant} and BiLLM~\cite{huang2024billm} for PTQ, spanning bit-widths from 1 to 8 bits. Our evaluation covers diverse language tasks using benchmarks such as Perplexity (WikiText2~\cite{merity2016pointer}, C4~\cite{raffel2020exploring}), 0-shot Commonsense Reasoning (PIQA~\cite{bisk2020piqa}, ARC-Easy/Challenge~\cite{clark2018think}, HellaSwag~\cite{zellers2019hellaswag}, Winogrande~\cite{sakaguchi2021winogrande}, BoolQ~\cite{boolq}), and 5-shot MMLU~\cite{hendrycks2020measuring}.
This study aims to: (1) benchmark quantization-induced performance trade-offs, (2) identify optimal methods for specific bit-widths, and (3) highlight unresolved challenges, particularly in ultra-low-bit regimes. We hope our findings will guide future research toward higher accuracy in compressed models, enhancing the practicality of Qwen3 and subsequent LLMs.

\section{Empirical Study}

\subsection{Experiment Settings}

We evaluate low-bit quantization across Qwen3's post-training models (0.6B, 1.8B, 4B, 7B, 14B, and 72B) as well as their pretraining versions (Qwen3-0.6/1.8/4/7/14B-Base), with pre-trained weights sourced from official repositories\textsuperscript{\ref{foot:qwen3}}.

\textbf{Quantization methods.} To comprehensively assess Qwen3's quantization robustness, we select 5 influential post-training quantization (PTQ) methods representing diverse technical approaches. All implementations adhere to their original open-source codebases\footnote{\ \url{https://github.com/IST-DASLab/gptq}, \url{https://github.com/mit-han-lab/llm-awq}, \url{https://github.com/mit-han-lab/smoothquant}, \url{https://github.com/Aaronhuang-778/BiLLM}}. Experiments were conducted on 1×NVIDIA A800 80GB GPU to ensure consistent evaluation conditions.

\textbf{Quantization protocol.}
To ensure fair comparison across all quantization methods, we maintain three key consistency measures: (1) all methods share identical calibration data (128 samples from C4 dataset~\cite{raffel2020exploring} with sequence length 2048), (2) for per-group quantization, channel-wise grouping adopts 128 block size following established practices in LLM quantization, and (3) weight-only quantization is applied uniformly from 1 to 8 bits. These controlled variables enable direct comparison of quantization method performance while minimizing confounding factors. Meanwhile, in weight-activation quantization methods, activations are quantized to 4 or 8 bits, which are the most commonly used settings, since lower bit-widths typically result in significant performance degradation.

\textbf{Evaluation protocol.}
For a comprehensive PTQ evaluation, we measure perplexity (PPL) on WikiText2~\cite{merity2016pointer} and a 256-sample subset of C4~\cite{raffel2020exploring} with a sequence length of 2048. Zero-shot accuracy is evaluated across six established reasoning benchmarks: PIQA~\cite{bisk2020piqa}, Winogrande~\cite{sakaguchi2021winogrande}, ARC-Easy and ARC-Challenge~\cite{clark2018think}, HellaSwag~\cite{zellers2019hellaswag}, and BoolQ~\cite{boolq}. Few-shot capability is further examined using 5-shot MMLU~\cite{hendrycks2020measuring}. This multi-dimensional evaluation framework provides a rigorous assessment of the quantized Qwen3's capabilities across various task types and difficulty levels.

\subsection{PTQ results}

\begin{figure}[t]
    \vspace{-0.10in}
    \centering
    \includegraphics[width=0.95\linewidth]{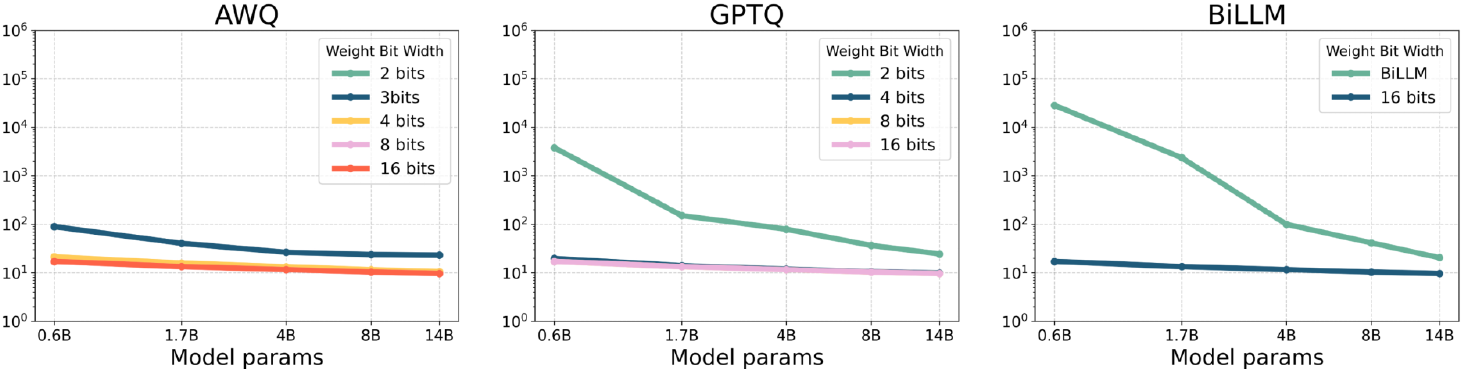}
    \vspace{-0.10in}
    \caption{PPL of per-group quantization on C4 of Qwen3-Base.} 
    \label{fig:qwen3_base_c4}
\end{figure}

\begin{figure}[t]
    \centering
    \includegraphics[width=0.95\linewidth]{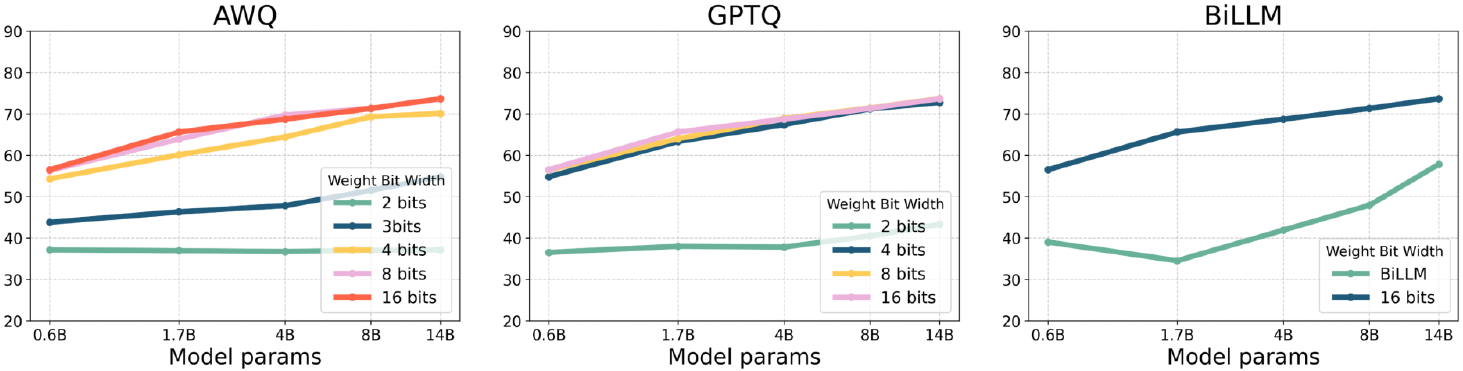}
    \vspace{-0.10in}
    \caption{0-shot Commonsense Reasoning Accuracy (Average of PIQA/Arc-Easy/Arc-Challenge/HellaSwag/Winogrande/BoolQ) of per-group quantization of Qwen3-Base.} 
    \label{fig:qwen3_base_zeroshot}
    \vspace{-0.10in}
\end{figure}

\begin{figure}[t]
    \vspace{-0.15in}
    \centering
    \includegraphics[width=0.82\linewidth]{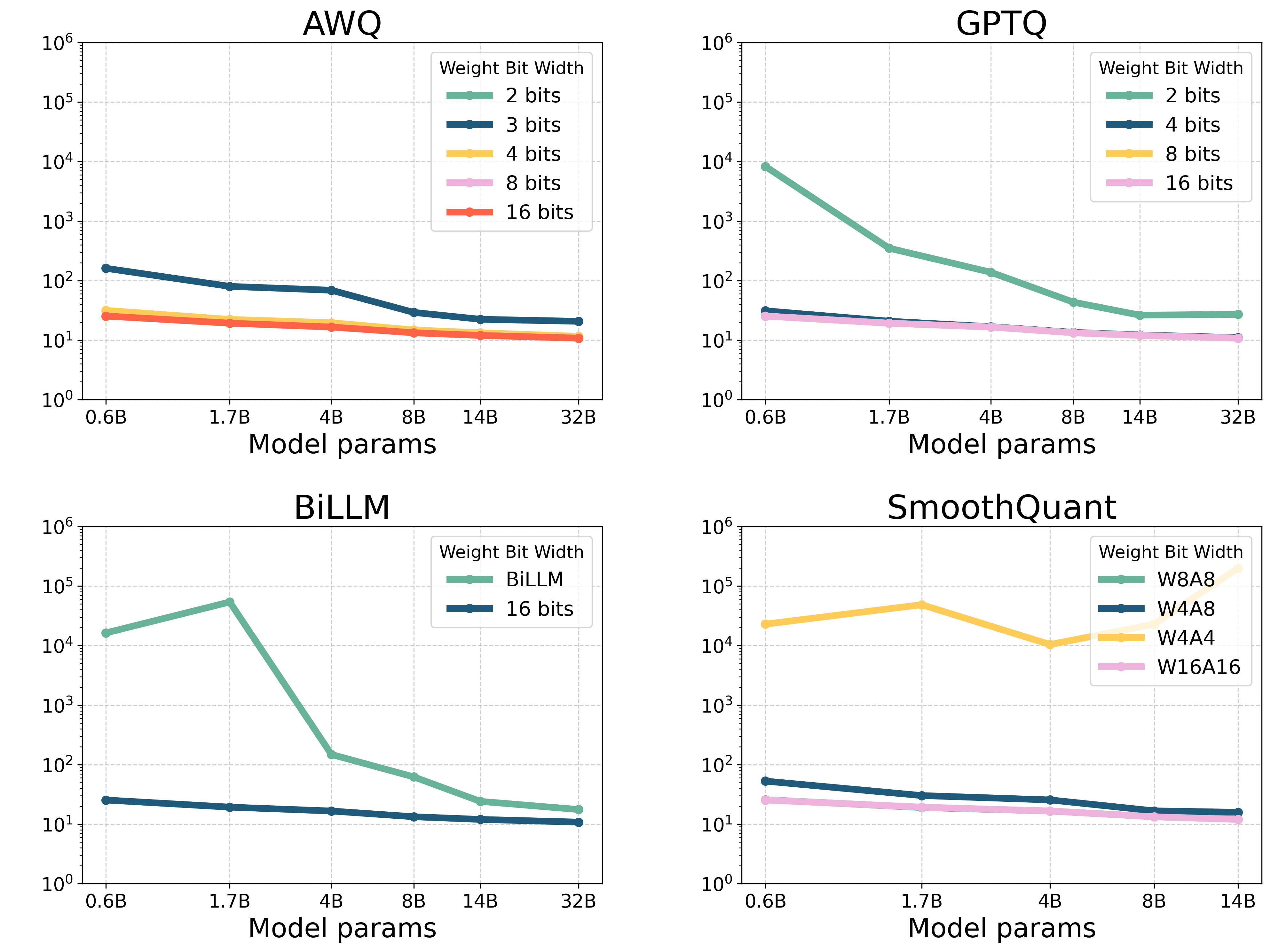}
    \vspace{-0.10in}
    \caption{PPL of per-group (AWQ, GPTQ, BiLLM) and per-channel (SmoothQuant) quantization methods on c4 of Qwen3.} 
    \label{fig:qwen3_c4}
\end{figure}

\begin{figure}[t]
    \centering
    \includegraphics[width=0.82\linewidth]{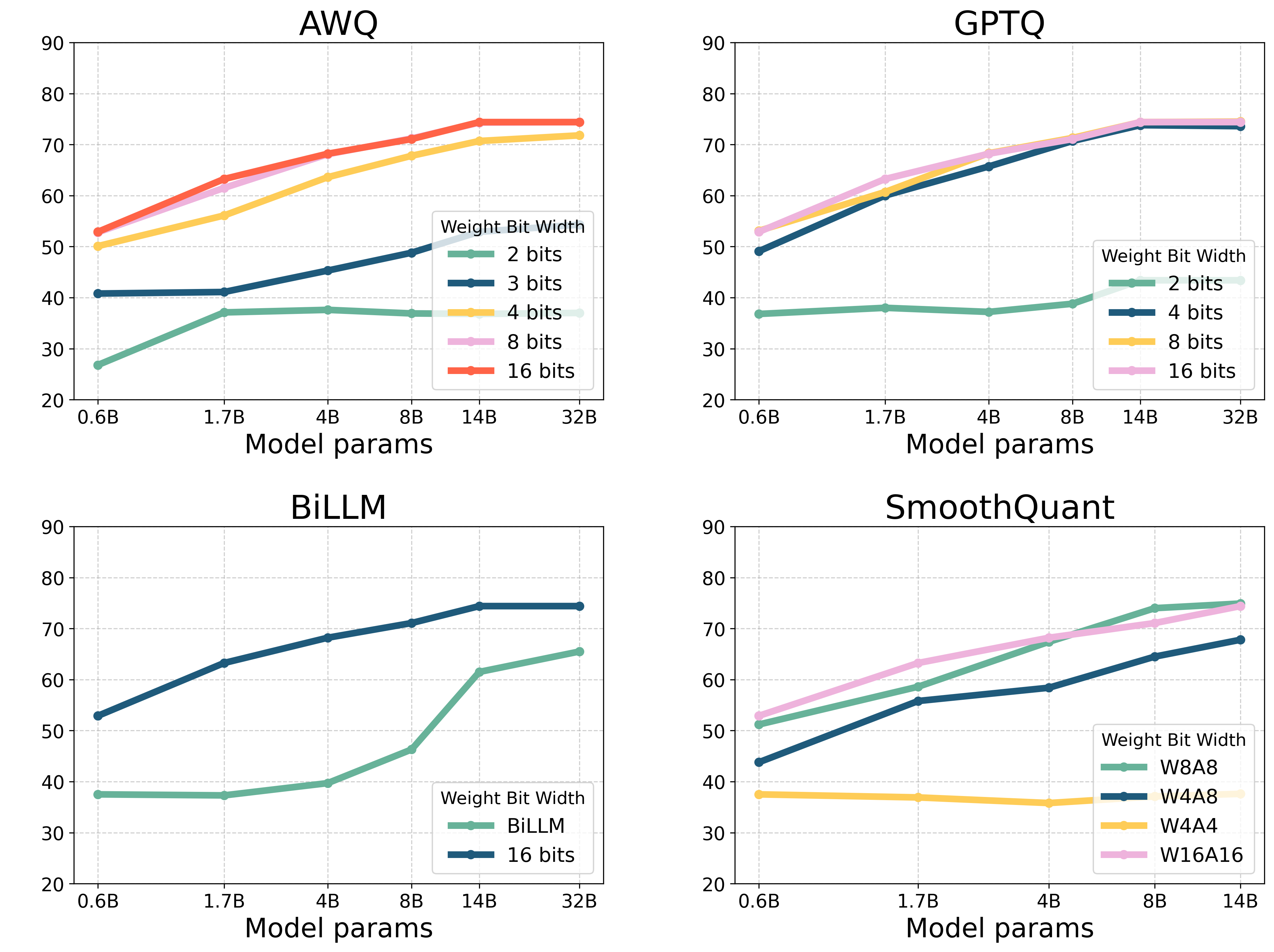}
    \vspace{-0.10in}
    \caption{0-shot Commonsense Reasoning Accuracy (Average of PIQA/Arc-Easy/Arc-Challenge/HellaSwag/Winogrande/BoolQ) per-group (AWQ, GPTQ, BiLLM) and per-channel (SmoothQuant) quantization methods on c4 of Qwen3.} 
    \label{fig:qwen3_zeroshot}
    \vspace{-0.10in}
\end{figure}

We present the detailed experimental results in Table~\ref{tab_qwen3-base-channel},Table~\ref{tab_qwen3-channel},Table~\ref{tab_qwen3-base-group} and Table~\ref{tab_qwen3-group}, and provide intuitive visual illustrations based on the data from Table~\ref{tab_qwen3-channel}, Table~\ref{tab_qwen3-base-group} and Table~\ref{tab_qwen3-group}, as shown in Figure~\ref{fig:qwen3_base_c4}, Figure~\ref{fig:qwen3_base_zeroshot}, Figure~\ref{fig:qwen3_c4} and Figure~\ref{fig:qwen3_zeroshot}.

\textbf{Impact of weight-only quantization.}
At 8 bits, Qwen3 consistently maintains near lossless performance, indicating that high-bit quantization still holds strong potential for practical deployment. However, when the bit-width is reduced to 4 bits, all quantization methods exhibit noticeable performance degradation. For example, Qwen-8B's MMLU score drops from 74.7 to 69.3, as shown in Table~\ref{tab_qwen3-group}. As the bit-width further decreases to 3 bits, although AWQ still retains some capability, most of the original model’s advantages are lost. At 2 bits, only methods like GPTQ, which leverage calibration-based compensation, manage to preserve a minimal level of performance. Meanwhile, we observe that Bi-LLM, a binarization method, demonstrates relatively promising results, surpassing even the 3-bit AWQ in the 32B model, highlighting the potential of binarization.

\textbf{Impact of activation quantization.} When applying SmoothQuant, one of the most classic activation quantization methods, we observe that even under the w8a8 setting, there is already a noticeable degradation in performance compared to the fp model. As the bit-width decreases to w4a8, the model suffers a significant performance drop, markedly worse than weight-only quantization. This result aligns with recent research findings, suggesting that large models may be particularly sensitive to activation quantization, possibly due to activation outliers, leading to substantial performance degradation.

\textbf{Comparison across varying parameter scales.}
We observe that larger models exhibit greater stability under quantization. Specifically, as shown in Table~\ref{tab_qwen3-group}, Qwen3-14B incurs only a 1\% drop in MMLU performance under 4-bit GPTQ compared to the full-precision model, whereas Qwen3-0.6B suffers a drop of around 10\% under the same setting, highlighting the ability of larger parameter spaces to mitigate quantization noise.

\textbf{Comparison with LLaMA3.} We previously conducted experiments on LLaMA3 using the classical methods~\cite{huang2024empirical}. Compared to the prior results on LLaMA3, Qwen3 exhibits more pronounced performance degradation under low-bit quantization (3 bits or fewer). Specifically, in LLaMA3-8B, AWQ with w3a16g128 quantization leads to a PPL increase on C4 from 9.2 to only 11.6, whereas in Qwen3-8B-Base, the same AWQ setting increases the PPL from 10.4 to 23.8. This aligns with our previous empirical observations and hypotheses: a more thorough pre-training process likely results in fewer redundant representations in stronger LLMs, making them more sensitive to quantization.


\begin{table*}[t]
\setlength{\tabcolsep}{5pt}
\small
\centering
\caption{2 to 8-bits per-channel PTQ results of Qwen3-Base Models. We report ppl on Wikitext2 and c4, 0-shot reasoning tasks and 5-shot mmlu performance. \#W denotes the weight quantization bit-width, and \#A denotes the activation quantization bit-width.}
\begin{adjustbox}{max width=\linewidth}
\begin{tabular}{ll||ccc|cc|ccccccc|cccccccccc}
\hline
\textbf{Model} & \textbf{Method} & \textbf{\#W} & \textbf{\#A} & \textbf{\#G} & \textbf{Wiki2$(\downarrow)$} & \textbf{c4$(\downarrow)$} & \textbf{PiQA} & \textbf{Arc E} & \textbf{Arc C} & \textbf{HellaS} & \textbf{WinoG} & \textbf{BoolQ} & \textbf{Avg$(\uparrow)$} & \textbf{MMLU} \\
\hline
\multirow{14}{*}{\centering 0.6B-Base}& FP16 & 16 & 16 &  / & 12.7 & 17.1 &70.0&65.6&33.9&41.1&58.5&69.7&56.5&52.3 \\
\cdashline{2-15}
& RTN & 8 & 16 & / &12.7	&17.1	&70.0	&65.8	&33.2	&41.1	&59.1	&70.2	&56.6	&52.4\\
& AWQ & 8 & 16 & / &12.7 &17.1 &70.3 & 65.0 & 33.4 &41.0 &59.2 &69.5 & 56.4&52.3 \\
& GPTQ & 8 & 16 & / & 12.7 & 17.1 & 70.1 & 65.9 & 33.4 & 40.9 & 58.3 & 70.0 & 56.4 & 52.2 \\
& SmoothQuant & 8 & 8 & / & 13.0 & 17.5 & 69.6 & 65.6 & 32.8 & 40.6 & 57.3 & 66.7 & 54.9 & 51.7\\
\cdashline{2-15}
& RTN & 4 & 16 & / &24.0	&30.8	&65.8	&58.9	&26.9	&35.7	&56.0	&63.3	&51.1	&35.9\\
& AWQ & 4 & 16 & / &15.6 & 20.1& 69.6 &  64.1 & 32.8  &  38.8&57.9 &65.3 &54.8 & 47.3\\
& GPTQ & 4 & 16 & / & 18.2 & 22.9 & 67.7 & 59.0 & 30.0 & 37.2 & 57.1 & 63.0 & 52.3 & 40.4 \\
& SmoothQuant & 4 & 8 & / & 31.5 & 39.1 & 63.5 & 53.6 & 27.3 & 34.3 & 52.9 & 63.1 & 45.7 & 32.8\\
& SmoothQuant & 4 & 4 & / & 5.25E4 & 4.21E4 & 51.6 & 26.9 & 21.6 & 25.2 & 48.5 & 51.0 & 37.8 & 23.9\\
\cdashline{2-15}
& AWQ & 3 & 16 & / &48.9 & 49.7& 62.9 & 50.1 & 25.2 & 31.7  & 53.2 & 62.1 &47.5 & 26.5\\
\cdashline{2-15}
& RTN & 2 & 16 & / &NaN	&NaN	&52.1	&24.7	&24.1	&25.9	&50.0	&44.3	&36.9	&23.8\\
& AWQ & 2 & 16 & / & 1.05E7&8.95E6 & 54.5 & 25.9 & 23.7 & 25.5 & 49.6 & 42.3 & 36.9 & 24.5\\
& GPTQ & 2 & 16 & / & 5.81E4 & NaN & 53.9 & 25.8 & 23.6 & 25.3 & 48.1 & 46.2 & 37.1 & 25.0 \\
\hline
\multirow{14}{*}{\centering 1.7B-Base} & FP16 & 16 & 16& / &9.39&13.4&75.7&73.2&41.5&49.2&64.2&79.2&65.6&61.0 \\
\cdashline{2-15}
& RTN & 8 & 16 & / &9.41	&13.4	&75.8	&74.0	&42.2	&49.3	&64.2	&79.5	&64.2	&60.5\\
& AWQ & 8 & 16 & / &9.39 & 13.4&75.8 &73.9 &41.6 &49.3 &64.2 &79.0 &64.0 &60.6 \\
& GPTQ & 8 & 16 & / & 9.39 & 13.4 & 75.7 & 73.8 & 41.6 & 49.3 & 64.2 & 78.7 &	63.9&	60.5\\
& SmoothQuant & 8 & 8 & / & 9.65	&13.8	&75	&73.8	&42.7	&48.8	&64	&77.6	&63.7	&60.2\\
\cdashline{2-15}
& RTN & 4 & 16 & / &16.8	&25.3	&71.2	&64.7	&34.4	&42.6	&59.9	&70.4	&57.2	&52.5\\
& AWQ & 4 & 16 & / &10.7 &14.8 &74.3 &69.1 &36.9 &47.3 &62.9 & 75.5& 61.0& 57.5\\
& GPTQ & 4 & 16 & / &11.0	&14.9	&72.0	&72.1	&40.4	&46.5	&63.3	&77.1	&61.9	&53.2\\
& SmoothQuant & 4 & 8 & / & 19.9 & 28.3 & 69.9 & 61.8 & 32.3 & 41.9 & 58.2 & 71.7 & 55.3 & 47.9\\
& SmoothQuant & 4 & 4 & / & 2.83E4 & 2.44E4 & 52.7 & 27.2 & 20.8 & 25.7 & 49.6 & 41.5 & 36.8 & 24.2\\

\cdashline{2-15}
& AWQ & 3 & 16 & / &17.8 & 22.3&68.4 & 61.4 &33.2 & 40.3 &  56.4&67.2 &54.5 & 40.5\\
\cdashline{2-15}
& RTN & 2 & 16 & / &1.96E7	&1.65E7	&51.4	&24.1	&23.4	&25.1	&50.9	&47.3	&37.0	&24.7\\
& AWQ & 2 & 16 & / & 3.35E6& 5.39E6&53.5 &25.1 &21.9 &25.5 &48.4 &43.2 & 36.3 & 24.9\\
& GPTQ & 2 & 16 & / &2.27E4	&9.49E3	&52.9	&24.0	&21.7	&25.8	&51.3	&45.1	&36.8	&24.5	\\
\hline
\multirow{14}{*}{\centering 4B-Base} & FP16 & 16 &16 & / &7.90&11.6&78.1&79.0&48.4&54.6&70.0&82.9&68.7&73.0\\
\cdashline{2-15}
& RTN & 8 & 16 & / &7.90	&11.6	&78.1	&79.0	&48.5	&54.5	&70.3	&82.9	&68.9	&72.8\\
& AWQ & 8 & 16 & / & 7.90 & 11.6& 78.0& 79.0& 47.9& 54.5& 70.6& 83.5& 68.9&72.8 \\
& GPTQ & 8 & 16 & / &7.89	&11.6	&78.0	&79.0	&48.3	&54.6	&70.6	&83.3	&69.0	&73.0\\
& SmoothQuant & 8 & 8 & / & 8.13 & 11.9 & 77.0 & 78.3 & 48.3 & 54.5 & 70.5 & 82.3 & 68.5
 & 72.0 \\
\cdashline{2-15}
& RTN & 4 & 16 & / &10.2	&14.4	&75.5	&73.0	&43.3	&51.6	&66.8	&73.9	&64.0	&65.0\\
& AWQ & 4 & 16 & / &8.79 &12.6 & 77.0 & 78.6 & 48.0 & 53.7 & 67.6 & 79.5 &  67.4 & 69.2\\
& GPTQ & 4 & 16 & / &8.70	&12.4	&76.6	&78.4	&46.9	&52.9	&68.2	&79.5	&67.1	&68.9\\
& SmoothQuant & 4 & 8 & / & 13.2 & 18.2 & 73.4 & 66.9 & 40.8 & 49.6 & 62.7 & 64.1 &59.6 & 63.2 \\
& SmoothQuant & 4 & 4 & / & 1.22E4 & 1.13E4 & 51.3 & 26.6 & 20.9 & 25.9 & 47.4 & 42.2 &35.7 & 23.6 \\
\cdashline{2-15}
& AWQ & 3 & 16 & / &15.0 & 17.7&71.7 &66.2 & 38.3 & 46.7 &59.7 &48.8 &55.2 &51.6 \\
\cdashline{2-15}
& RTN & 2 & 16 & / &2.8E7	&2.67E7	&52.9	&25.2	&20.4	&25.5	&51.1	&49.3	&37.4	&22.9\\
& AWQ & 2 & 16 & / &1.90E7 & 2.06E7& 52.3& 24.5& 23.0& 25.6& 51.6&48.7 &37.62 &25.3 \\
& GPTQ & 2 & 16 & / &2.01E4	&1.07E4	&51.0	&26.3	&21.7	&25.4	&50.7	&42.8	&36.3	&24.5\\
\hline
\multirow{14}{*}{\centering 8B-Base} & FP16 & 16 & 16 & / &6.99&10.4&79.3&82.1&52.6&58.9&72.1&82.9&71.3&76.7 \\
\cdashline{2-15}
& RTN & 8 & 16 & / &7.00	&10.4	&79.2	&81.8	&52.6	&58.9	&72.5	&83.0	&71.3	&76.6\\
& AWQ & 8 & 16 & / &6.99 & 10.4&79.1 &81.8 &53.3 &58.8 &72.4 &82.9 &71.4 & 76.6\\
& GPTQ & 8 & 16 & / &6.99	&10.4	&79.1	&81.6	&53.2	&58.9	&72.4	&83.0	&71.4	&76.6\\
& SmoothQuant & 8 & 8 & / & 7.43 & 11.1 & 78.4 & 80.9 & 52.7 & 58.7 & 77.5 & 82.5 & 71.9 & 75.5 \\
\cdashline{2-15}
& RTN & 4 & 16 & / &9.90	&14.5	&77.6	&78.5	&48.7	&56.0	&68.2	&71.4	&66.7	&70.2\\
& AWQ & 4 & 16 & / & 7.73 & 11.2& 77.5&81.2 &51.7 &57.5 &72.5 &83.0 & 70.6& 73.8\\
& GPTQ & 4 & 16 & / &7.63	&11.0	&78.3	&81.9	&52.1	&57.4	&68.7	&83.7	&70.4	&72.7\\
& SmoothQuant & 4 & 8 & / & 12.3 & 18.4 & 73.9 & 73.1 & 43.0 & 52.7 & 61.9 & 66.9 & 61.9 &61.6 \\
& SmoothQuant & 4 & 4 & / & 1.56E4 & 1.44E4 & 51.1 & 26.7 & 20.3 & 26.0 & 49.4 & 39.6 &35.5 & 24.8 \\
\cdashline{2-15}
& AWQ & 3 & 16 & / &11.4 & 14.4& 74.6 & 72.7& 40.6& 52.2& 59.1& 65.6& 60.8 & 57.7\\
\cdashline{2-15}
& RTN & 2 & 16 & / &9.32E6	&7.08E6	&51.8	&23.8	&22.4	&25.5	&52.1	&52.4	&38.0	&24.7\\
& AWQ & 2 & 16 & / & 1.34E7& 8.60E6& 52.6& 25.3&22.1 &25.5 &48.5 &44.2 & 36.4& 24.8\\
& GPTQ & 2 & 16 & / &4.75E3	&2.56E3	&50.8	&24.1	&22.6	&25.3	&50.0	&45.1	&36.3	&24.5\\
\hline
\multirow{14}{*}{\centering 14B-Base} & FP16 & 16 & 16 & / &6.38&9.68&80.5&83.5&55.6&61.7&74.1&86.5&73.7&80.7 \\
\cdashline{2-15}
& RTN & 8 & 16 & / &6.38	&9.69	&80.5	&83.6	&55.3	&61.8	&74.0	&86.6	&73.6	&80.6\\
& AWQ & 8 & 16 & / & 6.38& 9.69&80.5 &83.5 &55.9 &61.9 &74.2 &86.6 &73.8 & 80.7\\
& GPTQ & 8 & 16 & / &6.37	&9.69	&80.5	&83.5	&55.6	&61.8	&74.0	&86.8	&73.7	&80.7\\
& SmoothQuant & 8 & 8 & / &6.62	&10.0	&80.2	&82.8	&54.5	&61.4	&73.8	&86.2	&73.2	&79.6\\
\cdashline{2-15}
& RTN & 4 & 16 & / &14.0	&19.1	&76.3	&76.4	&51.0	&53.5	&65.4	&76.7	&66.6	&75.7\\
& AWQ & 4 & 16 & / &7.00 &10.3 & 79.9& 82.2& 54.2&60.6 &71.7 &83.1 &72.0 &78.7 \\
& GPTQ & 4 & 16 & / &7.11	&10.3	&79.7	&81.5	&54.0	&60.6	&73.2	&81.8	&71.8	&78.5\\
& SmoothQuant & 4 & 8 & / &39.6	&46.0	&72.4	&69.6	&43.4	&46.3	&63.2	&73.8	&61.5	&72.7\\
& SmoothQuant & 4 & 4 & / &1.54E6	&2.1E6	&52.7	&26.0	&20.7	&25.3	&50.0	&40.6	&35.9	&24.2\\
\cdashline{2-15}
& AWQ & 3 & 16 & / &9.95 &13.1 &76.0 &73.0 &44.6 &55.9 &62.7 &77.0 &64.9 & 70.8\\

\cdashline{2-15}
& RTN & 2 & 16 & / &5.3E6	&4.62E6	&53.9	&24.1	&21.5	&25.7	&50.5	&45.4	&36.8	&25.4\\
& AWQ & 2 & 16 & / &1.41E7& 1.19E7 & 53.4&24.7 &23.2&25.3 &50.0 &44.5 & 36.9 & 23.8 \\
& GPTQ & 2 & 16 & / &5.69E3	&1.14E3	&51.1	&26.0	&23.5	&24.8	&51.0	&42.3	&36.4	&24.4\\
\hline
\end{tabular}
\end{adjustbox}
\label{tab_qwen3-base-channel}
\end{table*}

\begin{table*}[t]
\setlength{\tabcolsep}{5pt}
\small
\centering
\caption{{2 to 8-bits per-channel PTQ results of Qwen3 Models}}
\begin{adjustbox}{max width=\linewidth}
\begin{tabular}{ll||ccc|cc|ccccccc|cccccccccc}
\hline
\textbf{Model} & \textbf{Method} & \textbf{\#W} & \textbf{\#A} & \textbf{\#G} & \textbf{Wiki2$(\downarrow)$} & \textbf{c4$(\downarrow)$} & \textbf{PiQA} & \textbf{Arc E} & \textbf{Arc C} & \textbf{HellaS} & \textbf{WinoG} & \textbf{BoolQ} & \textbf{Avg$(\uparrow)$} & \textbf{MMLU} \\
\hline
\multirow{14}{*}{\centering 0.6B}& FP16 & 16 & 16 & / &20.9&25.4&67.3&60.8&31.7&37.6&56.2&63.9&52.9&47.1\\
\cdashline{2-15}
& RTN & 8 & 16 & / &20.9	&25.4	&68.2	&61.1	&31.7	&37.6	&56.7	&63.1	&53.1	&47.0\\
& AWQ & 8 & 16 & / &20.9 & 25.4& 67.8& 60.4& 31.8&37.5 &56.0 &63.3 & 52.8&46.9 \\
& GPTQ & 8 & 16 & / &20.9	&25.4	&67.5	&60.7	&30.6	&37.4	&56.4	&64.5	&52.9	&47.0\\
& SmoothQuant & 8 & 8 & / & 21.3 & 25.7 & 66.8 & 59.4 & 31.2 & 37.2 & 52.3 & 60.1 & 51.2 & 46.3 \\
\cdashline{2-15}
& RTN & 4 & 16 & / &37.5	&42.4	&63.3	&49.8	&25.6	&33.3	&53.0	&62.6	&47.9	&37.3\\
& AWQ & 4 & 16 & / & 25.8
& 29.9 &65.1 &53.7 &27.7 &35.8 &56.1 & 61.3& 50.0& 43.1\\
& GPTQ & 4 & 16 & / &33.0	&37.3	&62.9	&48.3	&27.4	&34.5	&54.7	&59.1	&47.8	&40.0\\
& SmoothQuant & 4 & 8 & / & 49.5 & 52.8 & 61.6 & 43.9 & 23.7 & 31.6 & 52.2 & 50.0 & 43.8 & 30.8 \\
& SmoothQuant & 4 & 4 & / & 3.35E4 & 2.29E4 & 50.5 & 25.5 & 23.5 & 25.9 & 49.6 & 50.0 & 37.5 & 24.8 \\
\cdashline{2-15}
& AWQ & 3 & 16 & / & 80.4& 72.9& 60.0& 38.6& 21.6& 30.5& 50.9& 57.5& 43.2&26.4 \\
\cdashline{2-15}
& RTN & 2 & 16 & / &9.090E7	&5.66E7	&51.5	&23.7	&22.5	&25.5	&51.5	&45.8	&36.8	&24.4\\
& AWQ & 2 & 16 & / & 3.86E7& 3.43E7 &52.7 &23.6 &22.3 &25.5 &51.9 &50.4 &37.7 & 25.5\\
& GPTQ & 2 & 16 & / &7.19E5	&5.91E5	&51.9	&25.2	&21.4	&25.2	&49.5	&44.3	&36.2	&24.5\\
\hline
\multirow{14}{*}{\centering 1.7B} & FP16 & 16 &16& /&16.7&19.2&72.5&72.4&40.1&46.0&60.9&77.6&63.25&60.0 \\
\cdashline{2-15}
& RTN & 8 & 16 & / &16.7	&19.2	&72.3	&72.5	&39.3	&46.3	&61.7	&76.4	&61.4	&59.7\\
& AWQ & 8 & 16 & / &16.7 & 19.3 & 72.5& 72.1&39.7 &46.0 &61.7 &77.8 &61.6 & 60.0\\
& GPTQ & 8 & 16 & / &16.8	&19.3	&72.2	&72.6	&40.1	&46.2	&61.6	&77.0	&61.6	&59.9\\
& SmoothQuant & 8 & 8 & / & 16.4 & 19.0 & 71.3 & 69.9 & 39.3 & 45.9 & 58.6 & 75.5 & 58.6 & 58.9\\
\cdashline{2-15}
& RTN & 4 & 16 & / &28.7	&27.8	&68.4	&55.8	&32.7	&41.1	&55.6	&72.8	&54.4	&47.9\\
& AWQ & 4 & 16 & / &19.2 &21.2 & 71.5& 66.3& 37.1& 43.5& 56.7&75.3 &58.4 & 53.9\\
& GPTQ & 4 & 16 & / &21.0	&22.1	&68.2	&66.9	&35.9	&43.5	&60.0	&75.1	&58.3	&52.8\\
& SmoothQuant & 4 & 8 & / & 29.6 & 30.2 & 70.8 & 58.1 & 32.7 & 40.3 & 53.9 & 70.8 & 55.8 & 44.1\\
& SmoothQuant & 4 & 4 & / & 3.28E4 & 4.86E4 & 52.0 & 27.2 & 20.7 & 25.8 & 49.5 & 43.2 & 36.9 & 24.6\\
\cdashline{2-15}
& AWQ & 3 & 16 & / &29.0 & 32.5 & 63.5& 49.8 &24.7 &36.5 &52.6 &61.2 &48.1 & 35.6 \\
\cdashline{2-15}
& RTN & 2 & 16 & / &1.65E7	&1.63E7	&52.3	&24.5	&22.8	&25.6	&51.9	&48.4	&37.6	&24.4\\
& AWQ & 2 & 16 & / & 6.71E6 & 7.97E6 & 54.8&25.0 &22.4 &25.3 &50.4 &45.0& 37.2& 24.2\\
& GPTQ & 2 & 16 & / &1.43E5	&3.89E3	&53.7	&25.3	&21.9	&25.8	&50.7	&44.5	&37.0	&24.4\\
\hline
\multirow{14}{*}{\centering 4B} & FP16 & 16 & 16 & / &13.7&16.6&75.0&80.5&50.6&52.2&65.8&85.1&68.2&69.7\\
\cdashline{2-15}
& RTN & 8 & 16 & / &13.6	&16.6	&74.9	&80.3	&50.4	&52.2	&66.3	&85.0	&68.2	&69.8\\
& AWQ & 8 & 16 & / & 13.6& 16.6& 75.1& 80.7& 50.7&52.3 &66.4 &85.0 & 68.4& 69.5 \\
& GPTQ & 8 & 16 & / &16.8	&19.3	&72.2	&72.6	&40.1	&46.2	&61.6	&77.0	&61.6	&59.9\\
& SmoothQuant & 8 & 8 & / &13.7	&16.6	&73.9	&79.5	&49.5	&51.7	&64.7	&85.0	&67.4	&69.3\\
\cdashline{2-15}
& RTN & 4 & 16 & / &17.6	&20.3	&73.2	&75.3	&44.7	&48.6	&61.8	&81.6	&64.2	&63.0\\
& AWQ & 4 & 16 & / &16.6 & 18.9& 73.8&76.6 &46.2 &50.1 &63.1 &81.7 &65.3 & 66.0\\
& GPTQ & 4 & 16 & / &14.5	&17.5	&74.4	&76.6	&45.4	&50.1	&63.9	&82.7	&65.5	&65.8\\
& SmoothQuant & 4 & 8 & / &22.6	&25.5	&71.5	&69.5	&42.2	&46.1	&57.5	&63.8	&58.4	&59.2\\
& SmoothQuant & 4 & 4 & / & 9.91E3 & 1.04E4 & 51.6 & 25.9 & 22.6 & 25.6 & 47.9 & 41.0 & 35.8 & 24.6\\
\cdashline{2-15}
& AWQ & 3 & 16 & / &33.4& 29.9&66.4 &61.7 &32.6 &42.0 &54.6 &63.6 &53.5 &43.9 \\
\cdashline{2-15}
& RTN & 2 & 16 & / &9.31E6	&8.17E6	&52.0	&24.6	&22.0	&25.5	&49.9	&48.9	&37.1	&24.0\\
& AWQ & 2 & 16 & / &6.53E6 & 5.55E6 & 49.9&23.4 &22.1 &25.5 &51.9 &47.7 &36.8 & 24.7\\
& GPTQ & 2 & 16 & / &2.06E4	&1.25E4	&52.9	&80.1	&21.3	&25.4	&50.4	&41.8	&45.3	&24.6\\
\hline
\multirow{14}{*}{\centering 8B} & FP16 & 16 & 16 & /&9.71&13.3&76.4&83.5&55.5&57.1&68.0&86.5&71.08&74.7 \\
\cdashline{2-15}
& RTN & 8 & 16 & / &9.69	&13.3	&76.6	&83.5	&55.7	&57.1	&67.6	&86.5	&71.2	&74.6\\
& AWQ & 8 & 16 & / &9.73 & 13.3&80.2&84.3&58.7&50.9&72.8&89.4&72.7& 74.7\\
& GPTQ & 8 & 16 & / &9.70	&13.3	&76.7	&83.5	&55.5	&57.1	&68.4	&86.7	&71.3	&74.6\\
& SmoothQuant & 8 & 8 & / & 9.62 & 13.3 & 76.4 & 82.6 & 55.9 & 57.0 & 68.4 & 86.2 & 74.0 & 74.0\\
\cdashline{2-15}
& RTN & 4 & 16 & / &12.0	&15.6	&75.4	&79.1	&48.5	&53.8	&63.6	&78.9	&66.5	&68.2\\
& AWQ & 4 & 16 & / &10.5&14.2&79.4&83.1&57.2&59.7&72.0&88.4&73.3& 71.9 \\
& GPTQ & 4 & 16 & / &10.3	&13.8	&76.0	&80.1	&49.7	&55.6	&66.1	&84.5	&68.6	&71.6\\
& SmoothQuant & 4 & 8 & / & 12.5 & 16.7 & 73.4 & 71.0 & 44.3 & 51.7 & 62.1 & 79.1 & 64.5 & 63.2\\
& SmoothQuant & 4 & 4 & / & 3.36E4 & 2.29E4 & 50.5 & 25.5 & 25.7 & 26.7 & 52.2 & 39.6 & 37.1 & 24.8\\
\cdashline{2-15}
& AWQ & 3 & 16 & / &14.9 &18.5 &71.9&68.2&41.0&54.7&62.2&78.5&62.8&52.2 \\
\cdashline{2-15}
& RTN & 2 & 16 & / &NaN	&NaN	&52.9	&25.3	&22.3	&25.3	&51.2	&49.8	&37.8	&24.4\\
& AWQ & 2 & 16 & / &4.16E7 & 2.89E7&52.6&25.3&21.4&25.3&48.8&46.6&36.7&24.7 \\
& GPTQ & 2 & 16 & / &4.66E3	&2.93E3	&51.8	&25.8	&22.8	&25.5	&49.3	&41.8	&36.1	&24.2\\
\hline
\multirow{14}{*}{\centering 14B} & FP16 & 16 & 16&/ &8.64&12.0&80.0&84.3&59.0&60.9&72.9&89.4&74.4&78.5 \\
\cdashline{2-15}
& RTN & 8 & 16 & / &8.63	&12.0	&80.0	&84.2	&59.0	&60.9	&72.8	&89.3	&74.4	&78.5\\
& AWQ & 8 & 16 & / & 8.63& 12.0&76.8 &83.5 &55.2 & 57.1& 68.0& 86.7&71.2 &78.5 \\
& GPTQ & 8 & 16 & / &8.63	&12.0	&80.1	&84.2	&58.3	&61.0	&73.3	&89.1	&74.3	&78.4\\
& SmoothQuant & 8 & 8 & / & 8.69 & 12.1 & 78.6 & 83.5 & 57.9 & 61.1 & 73.1 & 89.2 & 74.9 & 77.8\\
\cdashline{2-15}
& RTN & 4 & 16 & / &9.98	&13.9	&77.9	&80.6	&51.8	&58.9	&68.7	&86.3	&70.7	&74.7\\
& AWQ & 4 & 16 & / &9.59 &13.1&76.0 &79.2 &51.7 &55.6 &66.1 &85.8 &69.1 & 76.3\\
& GPTQ & 4 & 16 & / &9.16	&12.6	&78.8	&81.9	&57.1	&59.6	&72.5	&87.8	&72.9	&75.9\\
\cdashline{2-15}
& AWQ & 3 & 16 & / &12.4&15.2 &72.8 &65.8 &37.8 &48.9 &56.1 &75.9 &59.6 & 67.1\\
& SmoothQuant & 4 & 8 & / & 11.0 & 15.8 & 75.8 & 79.0 & 50.5 & 57.0 & 66.3 & 82.2 & 67.8 & 71.3\\
& SmoothQuant & 4 & 4 & / & 2.16E5 & 1.99E5 & 51.2 & 25.8 & 26.5 & 25.8 & 50.2 & 38.5 & 37.6 & 24.9\\
\cdashline{2-15}
& RTN & 2 & 16 & / &2.05E6	&2.75E6	&53.5	&24.0	&23.9	&25.7	&48.5	&43.1	&36.4	&24.5\\
& AWQ & 2 & 16 & / &1.26E7 & 1.18E7 & 52.9&25.6 &24.3 &25.4 &53.4 &54.6 &39.4 & 25.0 \\
& GPTQ & 2 & 16 & / &2.28E3	&8.75E2	&52.2	&25.1	&23.4	&25.3	&49.4	&44.1	&36.6	&24.0\\
\hline
\multirow{11}{*}{\centering 32B}& FP16 & 16 & 16 & / &7.61&10.8&80.9&84.4&57.8&63.9&73.6&86.6&74.4&81.2 \\
\cdashline{2-15}
& RTN & 8 & 16 & / &7.60	&10.8	&80.8	&84.5	&58.2	&63.9	&73.0	&86.5	&74.5	&81.2\\
& AWQ & 8 & 16 & / & 7.60& 10.8&81.3&84.6&58.1&63.9&73.0&86.2&74.5&81.3\\
& GPTQ & 8 & 16 & / &7.60	&10.8	&81.1	&84.3	&57.5	&63.9	&72.5	&86.2	&74.3	&81.3\\
\cdashline{2-15}
& RTN & 4 & 16 & / &38.5	&35.0	&71.9	&73.0	&49.7	&44.4	&62.9	&85.8	&64.6	&78.4\\
& AWQ & 4 & 16 & / &8.21 & 11.3&79.8&82.5&56.9&63.1&71.8&84.5&73.1& 79.9\\
& GPTQ & 4 & 16 & / & 8.33	&11.4	&79.7	&82.6	&57.8	&62.7	&67.6	&84.5	&72.5	&79.1 \\
\cdashline{2-15}
& AWQ & 3 & 16 & /&12.3 &14.9&75.3&71.7&46.5&57.3&62.1&74.5&64.6&70.0\\
\cdashline{2-15}
& RTN & 2 & 16 & /&4.53E7	&5.3E7	&53.2	&24.7	&24.1	&25.5	&50.8	&50.7	&38.2	&25.5\\
& AWQ & 2 & 16 & / & 2.95E6& 2.56E6&52.6&24.8&22.8&25.4&51.9&51.1&38.1&23.6\\
& GPTQ & 2 & 16 & / & 1.07E4	&3.58E3	&52.6	&25.3	&22.1	&25.6	&50.4	&40.4	&36.1	&24.7\\
\hline
\end{tabular}
\end{adjustbox}
\label{tab_qwen3-channel}
\end{table*}

\begin{table*}[t]
\setlength{\tabcolsep}{5pt}
\small
\centering
\caption{1 to 8-bits per-group PTQ results of Qwen3-Base Models. We report ppl on Wikitext2 and c4, 0-shot reasoning tasks and 5-shot mmlu performance. \#W denotes the weight quantization bit-width, \#A denotes the activation quantization bit-width, and \#G denotes the group size.}
\begin{adjustbox}{max width=\linewidth}

\begin{tabular}{l||lccc|cc|ccccccc|ccccccccc|c}
\hline
\textbf{Model} & \textbf{Method} & \textbf{\#W} & \textbf{\#A} & \textbf{\#G} & \textbf{Wiki2$(\downarrow)$} & \textbf{c4$(\downarrow)$} & \textbf{PiQA} & \textbf{Arc E} & \textbf{Arc C} & \textbf{HellaS} & \textbf{WinoG} & \textbf{BoolQ} & \textbf{Avg$(\uparrow)$} & \textbf{MMLU} \\
\hline
\multirow{9}{*}{\centering 0.6B-Base} & FP16 & 16 & 16 &  / & 12.7 & 17.1 &70.0&65.6&33.9&41.1&58.5&69.7&56.5&52.3\\
\cdashline{2-15}
& AWQ & 8 & 16 & 128 & 12.7 & 17.1 & 70.2&65.0 & 33.2& 41.0&58.9 &69.4 & 56.3& 52.4\\
& GPTQ & 8 & 16 & 128 & 12.7 & 17.1 & 69.9 & 65.6 & 33.4 & 40.9 & 58.6 & 69.9 & 56.4 & 52.4 \\
\cdashline{2-15}
& AWQ & 4 & 16 & 128 & 16.6 & 21.3 & 68.3& 63.7&31.5 & 38.1 &58.2 &66.2 & 54.3& 43.8\\
& GPTQ & 4 & 16 & 128 & 14.9 & 19.7 & 69.5 & 61.8 & 32.4 & 39.0 & 56.8 & 69.2 & 54.8 & 47.2 \\
\cdashline{2-15}
& AWQ & 3 & 16 & 128 & 85.9 & 89.3 & 60.0& 43.6 & 22.1& 29.3& 51.1& 57.1& 43.8 & 25.9\\
\cdashline{2-15}
& AWQ & 2 & 16 & 128 & 6.42E7 & 9.85E7 & 53.2 & 24.8& 21.9& 25.7& 50.8& 46.3& 37.1 & 25.5 \\
& GPTQ & 2 & 16 & 128 & 7.5E3 & 3.8E3 & 52.0 & 24.8 & 22.8 & 26.2 & 51.6 & 41.4 & 36.5 & 24.8 \\
\cdashline{2-15}
& Bi-LLM & 1.06 & 16 & 128 & 8.64E4 & 2.85E4 & 53.6 & 28.9 & 24.4 & 26.4 & 51.5 & 47.6 & 39.0 & 26.9\\
\hline
\multirow{9}{*}{\centering 1.7B-Base} & FP16 & 16 & 16& / &9.39&13.4&75.7&73.2&41.5&49.2&64.2&79.2&65.6&61.0\\
\cdashline{2-15}
& AWQ & 8 & 16 & 128 &9.39 & 13.4& 75.7& 74.0& 41.4& 49.3& 64.3& 78.9&63.9 &60.7 \\
& GPTQ & 8 & 16 & 128 & 9.38 & 13.4 & 75.8 & 73.8 & 41.6 & 49.3 & 64.0 & 79.4 & 64.0 & 60.6 \\
\cdashline{2-15}
& AWQ & 4 & 16 & 128 & 11.4 & 15.8 & 72.9& 74.8 & 40.5 & 45.4&62.4 &64.5 & 60.1& 54.6\\
& GPTQ & 4 & 16 & 128 & 9.99 & 14.0 & 74.6 & 74.2 & 40.4 & 47.5 & 64.6 & 78.7 & 63.3 & 56.7 \\
\cdashline{2-15}
& AWQ & 3 & 16 & 128 & 41.8 & 40.4 &63.3 &49.8 &24.9 &34.4 &52.8 &52.4 &46.3 & 28.5 \\
\cdashline{2-15}
& AWQ & 2 & 16 & 128 & 1.13E7 & 1.02E7 &52.4& 24.0&21.7 &25.6 &49.4 &48.5 & 36.9& 25.2\\
& GPTQ & 2 & 16 & 128 & 2.23E2	&1.52E2	&54.3	&28.3	&18.9	&27.6	&49.3	&49.4	&38.0	&26.7\\
\cdashline{2-15}
& Bi-LLM & 1.04 & 16 & 128 & 3.26E3 & 2.35E3 & 52.2 & 28.6 & 23.6 & 26.5 & 50.3 & 45.7 & 34.5 & 23.2\\
\hline
\multirow{9}{*}{\centering 4B-Base} & FP16 & 16 &16 & / &7.90&11.6&78.1&79.0&48.4&54.6&70.0&82.9&68.7&73.0\\
\cdashline{2-15}
& AWQ & 8 & 16 & 128 &7.90 & 11.6&78.0 &78.8 & 48.0&54.5 & 70.3& 83.0& 69.7&72.8 \\
& GPTQ & 8 & 16 & 128 & 7.89 & 11.6 & 78.1 & 78.8 & 48.0 & 54.5 & 70.6 & 83.5 & 68.9 & 72.9 \\
\cdashline{2-15}
& AWQ & 4 & 16 & 128 & 9.39 &13.3 & 75.7& 73.3& 44.3 &52.3 &66.1 &74.6 & 64.4 & 66.7\\
& GPTQ & 4 & 16 & 128 & 8.19 & 11.9 & 78.1 & 77.9 & 47.8 & 53.8 & 70.1 & 76.9 & 67.4 & 70.9 \\
\cdashline{2-15}
& AWQ & 3 & 16 & 128 & 26.3& 26.4&  62.4& 46.1& 27.9& 40.1& 55.9&54.6 & 47.8 & 37.5 \\
\cdashline{2-15}
& AWQ & 2 & 16 & 128 & 7.53E6 & 5.94E6 &53.4 &24.7 & 22.2 &25.8 &47.5 &46.8 &36.7& 26.3\\
& GPTQ & 2 & 16 & 128 & 1.13E2 & 79.0 & 55.4 & 28.6 & 19.9 & 28.3 & 48.7 & 46.0 & 37.8 & 25.0 \\
\cdashline{2-15}
& Bi-LLM & 1.07 & 16 & 128 & 153  & 99.6  & 59.0 & 36.2 & 23.9 & 30.3 & 52.7 & 56.9 & 41.9 & 25.1\\
\hline
\multirow{9}{*}{\centering 8B-Base} & FP16 & 16 & 16 & / &6.99&10.4&79.3&82.1&52.6&58.9&72.1&82.9&71.3&76.7 \\
\cdashline{2-15}
& AWQ & 8 & 16 & 128 &6.99 & 10.4& 79.2& 81.8 & 53.2 & 58.8 & 72.3 & 82.8& 71.3 & 76.6\\
& GPTQ & 8 & 16 & 128 & 6.99 & 10.4 & 79.3 & 81.7 & 52.9 & 58.9 & 72.8 & 82.8 & 71.4 & 76.7 \\
\cdashline{2-15}
& AWQ & 4 & 16 & 128 &8.11 &11.6 &77.9 &81.1 &50.3 & 56.7& 69.2& 80.7& 69.3 & 72.4\\
& GPTQ & 4 & 16 & 128 & 7.22	&10.6	&78.9	&82.6	&53.8	&58.5	&71.0	&82.4	&71.2	&75.4\\
\cdashline{2-15}
& AWQ & 3 & 16 & 128 & 22.6 & 23.8& 67.7& 53.2&30.4 &43.3 &54.0 & 60.1& 51.5& 36.6\\
\cdashline{2-15}
& AWQ & 2 & 16 & 128 & 1.66E7& 1.31E7 & 52.6 & 26.6 &22.6 & 25.5& 50.0 & 44.6 & 37.0 & 25.3 \\
& GPTQ & 2 & 16 & 128 & 53.1 & 36.6 & 57.4 & 30.6 & 20.5 & 33.4 & 52.5 & 48.7 & 40.5 & 26.8 \\
\cdashline{2-15}
& Bi-LLM & 1.05 & 16 & 128 & 48.2 & 41.4 & 62.5 & 43.8 & 25.9 & 34.3 & 53.0 & 63.9 & 47.9 & 30.1\\
\hline
\multirow{9}{*}{\centering 14B-Base} & FP16 & 16 & 16 & / &6.38&9.68&80.5&83.5&55.6&61.7&74.1&86.5&73.65&80.7\\
\cdashline{2-15}
& AWQ & 8 & 16 & 128 & 6.38 & 9.69 & 80.6 & 83.3&55.7 &61.9 &74.3 &86.7 &73.7 &80.7 \\
& GPTQ & 8 & 16 & 128 &6.37	&9.68	&80.6	&83.5	&56.0	&61.8	&73.7	&86.8	&73.7	&80.7\\
\cdashline{2-15}
& AWQ & 4 & 16 & 128 & 7.35 & 10.7 &78.4 &80.9 &52.8 & 59.3 & 72.2& 77.1& 70.1& 77.1 \\
& GPTQ & 4 & 16 & 128 &6.65	&9.90	&80.6	&82.4	&52.6	&61.3	&73.7	&85.7	&72.7	&79.8\\

\cdashline{2-15}
& AWQ & 3 & 16 & 128 &19.2 & 23.1& 69.7& 59.9& 36.9&47.6 &53.7 & 60.7& 54.8 & 43.7\\
\cdashline{2-15}
& AWQ & 2 & 16 & 128 & 2.68E7 &2.18E7 & 53.0 & 24.6 &23.0 & 25.3& 50.7 & 46.2 & 37.1 & 24.4\\
& GPTQ & 2 & 16 & 128 &27.9	&24.5	&59.5	&37.4	&22.2	&36.8	&51.5	&52.5	&43.3	&27.2\\
\cdashline{2-15}
& Bi-LLM & 1.05 & 16 & 128 & 20.9  & 20.4 & 69.7 & 61.2 & 35.0 & 41.8 & 65.2 & 73.5 & 57.8 & 39.9\\
\hline
\end{tabular}
\end{adjustbox}
\label{tab_qwen3-base-group}
\end{table*}

\begin{table*}[t]
\small
\centering
\caption{{1 to 8-bits per-group PTQ results of Qwen3 Models}}
\begin{adjustbox}{max width=\linewidth}
\begin{tabular}{ll||ccc|cc|ccccccc|cccccccccc}
\hline
\textbf{Model} & \textbf{Method} & \textbf{\#W} & \textbf{\#A} & \textbf{\#G} & \textbf{Wiki2$(\downarrow)$} & \textbf{c4$(\downarrow)$} & \textbf{PiQA} & \textbf{Arc E} & \textbf{Arc C} & \textbf{HellaS} & \textbf{WinoG} & \textbf{BoolQ} & \textbf{Avg$(\uparrow)$} & \textbf{MMLU} \\
\hline
\multirow{9}{*}{\centering 0.6B} & FP16 & 16 & 16 & / &20.9&25.4&67.3&60.8&31.7&37.6&56.2&63.9&52.9&47.1\\
\cdashline{2-15}
& AWQ & 8 & 16 & 128 &20.9 &25.4 & 67.6&60.6 & 31.5&37.5 &56.4 & 63.2& 52.8& 47.0\\
& GPTQ & 8 & 16 & 128 & 20.9 & 25.4 & 67.6 & 60.9 & 30.9 & 37.6 & 56.6 & 65.2 & 53.1 & 47.0 \\
\cdashline{2-15}
& AWQ & 4 & 16 & 128 & 26.9& 31.6 & 64.3 & 54.1& 25.8 & 35.1 & 55.4 & 65.6& 50.1& 42.1 \\
& GPTQ & 4 & 16 & 128 & 25.3 & 31.1 & 64.7 & 48.1 & 29.4 & 35.7 & 53.4 & 63.6 & 49.1 & 44.0 \\
\cdashline{2-15}
& AWQ & 3 & 16 & 128 & 2.2E2 & 1.61E2&56.3 &33.9 & 21.7& 28.2&50.0 &54.6 &40.8 & 25.0 \\
\cdashline{2-15}
& AWQ & 2 & 16 & 128 & 1.21E7& 8.82E6 &53.9 &24.3 & 22.2 & 25.5 & 51.6&43.4 & 36.8& 25.7\\
& GPTQ & 2 & 16 & 128 & 2.38E4 & 8.27E3 & 53.2 & 26.2 & 20.9 & 25.6 & 51.1 & 43.9 & 36.8 & 23.5 \\
\cdashline{2-15}
& Bi-LLM & 1.06  & 16 & 128 & 5.87E4 & 1.63E4 & 53.1 & 27.2 & 25.2 & 26.2 & 50.6 & 42.6 & 37.5 & 23.3\\
\hline
\multirow{9}{*}{\centering 1.7B} & FP16 & 16 &16& /&16.7&19.3&72.5&72.4&40.1&46.0&60.9&77.6&63.25&60.0\\
\cdashline{2-15}
& AWQ & 8 & 16 & 128 & 16.8& 19.2& 72.6&71.9 &39.2 &46.1 &61.6 &77.7 &61.5 & 59.8\\
& GPTQ & 8 & 16 & 128 & 16.7 & 19.2 & 72.1 & 72.3 & 39.8 & 46.2 & 56.6 & 77.1 & 60.7 & 59.9 \\
\cdashline{2-15}
& AWQ & 4 & 16 & 128 &19.6 &22.3 & 69.4 & 63.1 & 33.5 & 42.3 & 58.4 &69.8 &  56.1& 52.5\\
& GPTQ & 4 & 16 & 128 & 19.5 & 20.8 & 70.6 & 71.2 & 38.2 & 44.0 & 59.3 & 76.8 & 60.0 & 55.7 \\
\cdashline{2-15}
& AWQ & 3 & 16 & 128 & 84.0 & 80.0& 58.4& 36.5& 22.2& 29.7& 50.2& 49.6& 41.1& 26.7 \\
\cdashline{2-15}
& AWQ & 2 & 16 & 128 & 7.52E6 & 7.66E6&53.4 &24.7 &21.0 &25.9 &51.6 &46.0 & 37.1& 25.2\\
& GPTQ & 2 & 16 & 128 & 6.55E2 & 3.5E2 & 53.2 & 25.7 & 20.1 & 26.3 & 52.0 & 50.3 & 38.0 & 25.6 \\
\cdashline{2-15}
& Bi-LLM & 1.04  & 16 & 128 & 7.54E4 & 5.40E4 & 52.7 & 28.5 & 24.7 & 26.6 & 48.9 & 42.4 & 37.3 & 25.1\\
\hline
\multirow{9}{*}{\centering 4B} & FP16 & 16 & 16 & / &13.7&16.6&75.0&80.5&50.6&52.2&65.8&85.1&68.2&69.7 \\
\cdashline{2-15}
& AWQ & 8 & 16 & 128 & 13.6& 16.6&74.9 &80.6 &50.5 &52.3 &65.6 &85.1 &68.1 & 69.6\\
& GPTQ & 8 & 16 & 128 & 13.6 & 16.6 & 75.0 & 80.5 & 50.8 & 52.3 & 66.0 & 85.1 & 68.3 & 69.7 \\
\cdashline{2-15}
& AWQ & 4 & 16 & 128 & 18.1& 19.6& 72.7 & 73.0 & 45.0 & 48.6  & 60.1 & 82.0 & 63.6& 64.1\\
& GPTQ & 4 & 16 & 128 & 13.5 & 16.8 & 74.0 & 71.2 & 48.9 & 50.9 & 65.0 & 84.3 & 65.7 & 67.6 \\
\cdashline{2-15}
& AWQ & 3 & 16 & 128 & 91.0 & 68.9 &61.7 & 42.4  &  27.2 & 35.5 & 51.5 &  55.8 &45.3 &31.5 \\
\cdashline{2-15}
& AWQ & 2 & 16 & 128 &1.38E7 & 1.44E7&52.6  & 26.7  & 22.1 & 25.4 & 48.8 & 49.7 &  37.6 & 24.6\\
& GPTQ & 2 & 16 & 128 & 1.95E2 & 1.38E2 & 53.6 & 26.3 & 21.1 & 27.0 & 51.4 & 44.1 & 37.2 & 24.4 \\
\cdashline{2-15}
& Bi-LLM & 1.07 & 16 & 128 & 285 & 148 & 56.7 & 30.6 & 22.8 & 32.9 & 48.9 & 46.2 & 39.7 & 26.3\\
\hline
\multirow{9}{*}{\centering 8B} & FP16 & 16 & 16 & /&9.71&13.3&76.4&83.5&55.5&57.1&68.0&86.5&71.08&74.7\\
\cdashline{2-15}
& AWQ & 8 & 16 & 128 & 9.72&13.3 &76.8 &83.5 &55.5 &57.1 &67.5 &86.5 &71.2 &74.5 \\
& GPTQ & 8 & 16 & 128 & 9.70 & 13.3 & 76.7 & 83.7 & 55.6 & 57.1 & 67.9 & 86.7 & 71.3 & 74.7 \\
\cdashline{2-15}
& AWQ & 4 & 16 & 128 & 11.3 &14.8 & 75.5 & 80.8 & 48.8 & 54.6 & 64.7 & 82.1 & 67.8 & 69.3\\
& GPTQ & 4 & 16 & 128 & 9.96 & 13.5 & 76.0 & 82.4 & 54.2 & 56.6 & 68.4 & 86.5 & 70.7 & 73.4 \\
\cdashline{2-15}
& AWQ & 3 & 16 & 128 &27.5 & 29.2& 65.2 & 45.2 & 26.6 & 40.3 &  54.1 & 61.6 & 48.8& 33.2\\
\cdashline{2-15}
& AWQ & 2 & 16 & 128 &1.21E7 & 8.14E6 & 51.4 & 25.8 & 23.4 & 25.3 & 51.7 & 43.7 &36.9 & 25.3\\
& GPTQ & 2 & 16 & 128 & 52.1 & 43.5 & 55.5 & 27.9 & 20.6 & 30.3 & 51.5 & 46.9 & 38.8 & 25.0 \\
\cdashline{2-15}
& Bi-LLM & 1.05 & 16 & 128 & 90.4   & 62.1   & 60.3 & 38.2 & 24.5 & 38.0 & 53.5 & 63.1 & 46.3 & 32.8\\
\hline
\multirow{9}{*}{\centering 14B} & FP16 & 16 & 16& /&8.64&12.0&80.0&84.3&59.0&60.9&72.9&89.4&74.4&78.5\\
\cdashline{2-15}
& AWQ & 8 & 16 & 128 & 8.63& 12.0&80.0 &84.3 &58.4 &60.9 &72.8 &89.5 & 74.3&78.5\\
& GPTQ & 8 & 16 & 128 & 8.63 & 12.0 & 80.1 & 84.4 & 58.4 & 60.9 & 73.2 & 89.3 & 74.4 & 78.4 \\
\cdashline{2-15}
& AWQ & 4 & 16 & 128 &9.48 & 13.3& 77.3 & 79.5 &  51.9 &  58.4 &70.0  & 87.0 & 70.7 & 75.9\\
& GPTQ & 4 & 16 & 128 & 8.88 & 12.2 & 79.1 & 83.8 & 57.9 & 60.5 & 72.0 & 89.2 & 73.8 & 77.4 \\
\cdashline{2-15}
& AWQ & 3 & 16 & 128 &19.4 &22.4 &68.2 & 52.2 & 29.8 &  46.1 &  54.4&66.4  & 52.9 &47.7\\
\cdashline{2-15}
& AWQ & 2 & 16 & 128 &6.06E6 &5.24E6& 52.3 & 24.9 & 23.1 & 25.5 & 48.3 & 46.6 &36.8 & 25.0\\
& GPTQ & 2 & 16 & 128 & 25.5 & 26.3 & 60.4 & 36.0 & 22.0 & 36.0 & 53.4 & 52.8 & 43.4 & 28.5 \\
\cdashline{2-15}
& Bi-LLM & 1.05 & 16 & 128 & 29.1 & 24.1 & 69.4 & 59.3 & 35.9 & 56.4 & 65.7 & 82.1 & 61.5 & 39.9\\
\hline
\multirow{9}{*}{\centering 32B} & FP16 & 16 & 16 & / &7.61&10.8&80.9&84.4&57.8&63.9&73.6&86.6&74.4&81.2 \\
\cdashline{2-15}
& AWQ & 8 & 16 & 128 &7.60&10.8&80.8&84.3&57.9&63.9&72.9&86.4& 74.4 &81.3 \\
& GPTQ & 8 & 16 & 128 & 7.61	&10.8	&80.8	&84.4	&58.2	&63.9	&73.4	&86.4	&74.5	&81.4\\
\cdashline{2-15}
& AWQ & 4 & 16 & 128 & 8.55 & 11.6&79.1&81.3&54.2&61.9&68.0&86.4&71.8& 78.0\\
& GPTQ & 4 & 16 & 128 & 7.86	&11.0	&79.8	&83.8	&57.6	&63.3	&71.5	&85.7	&73.6	&80.6 \\
\cdashline{2-15}
& AWQ & 3 & 16 & 128 & 19.1& 20.7&71.0 &61.6&35.6&42.7&53.1&61.8&54.3& 54.9\\
\cdashline{2-15}
& AWQ & 2 & 16 & 128 & NaN& NaN&54.8&26.0&21.7&25.7&49.3&44.3&37.0&24.6 \\
& GPTQ & 2 & 16 & 128 &38.4	&27.1	&60.6	&33.1	&25.6	&38.6	&51.9	&50.8	&43.4	&28.1
 \\
\cdashline{2-15}
& Bi-LLM & 1.06 & 16 & 128 & 17.1 & 17.6 & 73.2 & 63.8 & 43.3 & 67.8 & 66.7 & 78.1 & 65.5 & 57.5\\
\hline
\end{tabular}
\end{adjustbox}
\label{tab_qwen3-group}
\end{table*}

\section{Conclusion}
The newly released Qwen3 series has emerged as one of the most capable open-source LLM families, garnering substantial attention from both academia and industry.  In this study, we present the first systematic evaluation of Qwen3's robustness under various low-bit quantization schemes, with particular focus on post-training quantization methods.  Our investigation seeks to establish practical boundaries for deploying Qwen3 in resource-constrained scenarios through comprehensive quantization analysis.

Our experimental results reveal that while Qwen3 maintains competitive performance at higher bit-widths (4-bit and above), it exhibits more pronounced performance degradation compared to previous model generations when quantized to 3-bit or below.  This observation aligns with our hypothesis that advanced pre-training techniques, which Qwen3 extensively employs, tend to produce models with less parameter redundancy, consequently making them more sensitive to quantization-induced information loss. Notably, the performance drop becomes particularly significant in complex reasoning tasks and few-shot learning scenarios.

These findings underscore two critical implications: (1) current quantization techniques require further innovation to better preserve Qwen3's advanced capabilities, and (2) the trade-offs between model compression and performance retention need careful reconsideration for state-of-the-art LLMs. We believe our empirical analysis provides valuable guidance for future research directions in LLM quantization, particularly in developing methods that can maintain high accuracy at ultra-low bit-widths. As the field progresses, we anticipate these insights will contribute to more efficient deployment of powerful models like Qwen3, ultimately advancing the practical applications of large language models while reducing their computational overhead.

\textbf{Future Work.} We plan to evaluate more advanced forms of quantization methods, such as channel reordering-based approaches~\cite{yuan2023rptq} and rotation-based quantization strategies~\cite{liu2024spinquant}, to assess Qwen3's performance under these techniques, particularly regarding their impact on activation quantization.

\clearpage

\bibliographystyle{splncs04}
\bibliography{main}
\end{document}